\documentclass[conference]{IEEEtran}

\usepackage{amssymb,graphicx,url}
\usepackage{fancyhdr}
\usepackage[margin=1in]{geometry}

\usepackage{amsmath}

\ifCLASSINFOpdf
\else
\fi
\begin{document}

	\title{\LARGE{{Domain Adaptation for Robot Predictive Maintenance Systems}}}
	
	
	\author{\IEEEauthorblockN{Arash Golibagh Mahyari\textsuperscript{\textdagger}} \\
		\IEEEauthorblockA{Institute for Human and Machine Cognition (IHMC)\\
			Pensacola, FL\\
			Email: amahyari@ihmc.us}
			\and
			\IEEEauthorblockN{Thomas Locker} \\
		\IEEEauthorblockA{ABB Future Labs \\
			Baden-D\"attwil, Switzerland\\
			Email: thomas.locher@ch.abb.com}
			}

	
	\maketitle

	\begin{abstract}
		
	Industrial robots play an increasingly important role in a growing number of fields. For example, robotics is used to increase productivity while reducing costs in various aspects of manufacturing. Since robots are often set up in production lines, the breakdown of a single robot has a negative impact on the entire process, in the worst case bringing the whole line to a halt until the issue is resolved, leading to substantial financial losses due to the unforeseen downtime. Therefore, predictive maintenance systems based on the internal signals of robots have gained attention as an essential component of robotics service offerings. In the state of the art in predictive maintenance algorithms, features are extracted from the internal signals of the robot while the robot is healthy in order to build a model representing normal robot behavior. Extracted features during operation are then compared to the model of normal behavior to detect any change in the robot's mechanical condition, which may indicate a (future) problem with the robot. The main shortcoming of existing predictive maintenance algorithms is that the extracted features typically differ significantly from the learnt model when the operation of the robot changes, incurring false alarms. In order to mitigate this problem, predictive maintenance algorithms require the model to be retrained with normal data of the new operation. In this paper, we propose a novel solution based on transfer learning to pass the knowledge of the trained model from one operation to another in order to prevent the need for retraining and to eliminate such false alarms. The deployment of the proposed unsupervised transfer learning algorithm on real-world datasets demonstrates that the algorithm can not only distinguish between operation and mechanical condition change, it further yields a sharper deviation from the trained model in case of a mechanical condition change and thus detects mechanical issues with higher confidence.

	\end{abstract}

	{\it{\bf keywords- predictive maintenance, Anomaly Detection, Predictive Maintenance, Transfer Learning, Domain Adaptation}}
	
	%
	
	\pagestyle{fancy}
	\renewcommand{\headrulewidth}{0pt} 

	\IEEEpeerreviewmaketitle

	\section{Introduction}
	
	\let\thefootnote\relax\footnotetext{\noindent \textsuperscript{*}Parts of this paper are patent pending. \\ \textsuperscript{\textdagger}This work was done while the author worked at the ABB Robotics R\&D Center, 3055 Orchard Dr., San Jose, CA 95134.}

	Robots have revolutionized the manufacturing process by performing tasks more efficiently and accurately at a lower operational cost. However, sudden breakdowns or malfunctions of robots can result in a steep decrease in production quality and quantity, which often entails substantial financial losses.
Robot failures and malfunctions can have various causes, e.g., moving heavy objects continuously leads to the deterioration of robots gears over time. As a result of this deterioration, gaps can appear between the gear teeth, which results in so-called \emph{backlash}. Backlash is a clearance or lost motion in a mechanism. It is noticeable in a gearbox when the direction of movement is reversed and the slack or lost motion is taken up before the reversal of motion is complete.

In order to prevent unscheduled maintenance due to sudden disruptions of normal operation, \emph{predictive maintenance} has gained a lot of attention recently~\cite{hornung2014model, liu2005model, mobley2002introduction, garcia2006simap}.
The goal of predictive maintenance is to prevent unexpected equipment failure by constantly monitoring the 
performance and condition of equipment in operation and to extrapolate from the current condition when corrective maintenance will be required \cite{grall2002continuous, hashemian2010state}. Predictive maintenance can be performed on a software and hardware level. Software faults are, e.g., communication problems, controller software malfunctioning, etc. Hardware issues are mostly related to broken sensors, broken gears, backlash, etc. The focus of this paper is to identify hardware issues through analyzing the internal signals of the robots. An accurate system capable of detecting hardware problems requires additional internal sensors. However, using extra sensors to perform predictive maintenance is not feasible in many situations because it increases the cost, complexity, system weight, and requires extra space inside the robot system~\cite{hornung2014model}. For this reason, predictive maintenance systems based on machine learning algorithms have gained popularity recently and this approach is the focus of this paper.
	
\begin{figure*}[th]
\begin{minipage}[b]{\linewidth}
  \centering
  \centerline{\includegraphics[width=8cm]{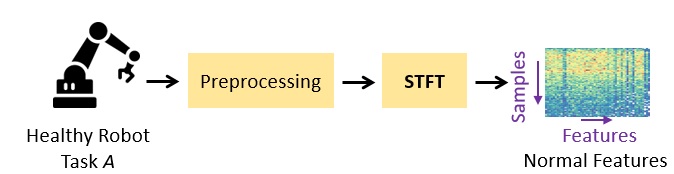}}

  \centerline{   (a)}\medskip
  \vspace{-2mm}
\end{minipage}

\begin{minipage}[b]{\linewidth}
  \centering
  \centerline{\includegraphics[width=12cm]{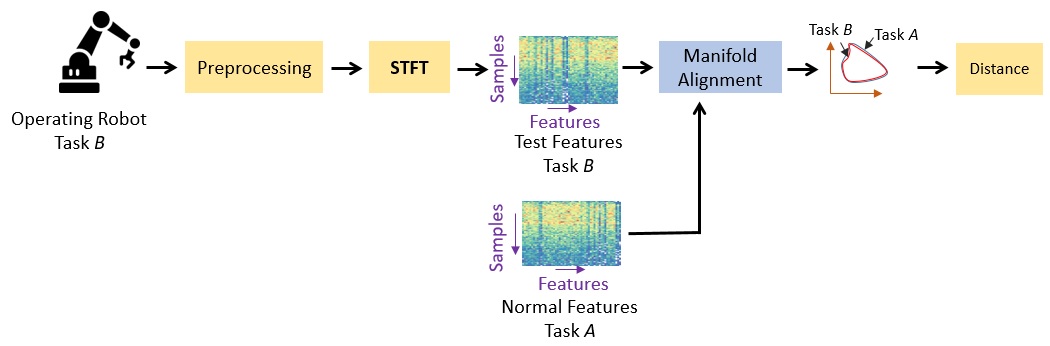}}

  \centerline{   (b)}\medskip
  \vspace{-4mm}
\end{minipage}

\caption{Flowchart of the proposed work: (a) Training from the healthy robot with task $A$; (b) Evaluating (test) the operating robot in the customer facility with unknown task $B$ using domain adaptation.}
\label{fig:schematic}
\vspace{-5mm}
\end{figure*}  	

There have been several research activities recently in this domain~\cite{hornung2014model, park2016multimodal, borgi2017data, sjostrand2012dynamic, pinto2019robot, moisescu2018retracted, sathish2019data}. Hornung et al.\ propose to map data to a positive and a negative group, where the positive group includes all areas of data space accessed during normal operation while data in the negative group belongs to the unknown area of data space~\cite{hornung2014model}. In order to approximate the positive data space, radial basis function kernels are used. Anomalies are detected by training a support vector machine (SVM) on positive and negative data. In another work, Park et al.\ propose to use multimodal sensory data such as haptic, auditory, visual, and kinematic signals to train a hidden Markov model (HMM)\cite{park2016multimodal}. The trained HMM provides a probability of the test data belonging to class associated with normal behavior, which can be used as a criterion for anomaly detection.
	
Generally, these approaches consist of data pre-processing, feature extraction, dimension reduction, and a model-based classifier. A model is trained for each axis of the robot based on the training data collected while the robot is healthy and performing a specific task. Later, the health of the robot is evaluated by collecting data and comparing it with the trained model.
	
The problem with existing predictive maintenance algorithms is that they raise false alarms when the task for which data is collected differs from the task used during the training phase. When the task changes, features such as frequency components extracted from data change as well. This change  in the features fools the systems to erroneously raise an alarm, although the robot is healthy and operating normally. In order to avoid false alarms due to an operational change, current systems require the model to be retrained when altering the task of the robot.
	
In this paper, we propose to use \emph{transfer learning}, also known as knowledge transfer and domain adaptation, as an intermediate step in order to avoid retraining the model every time the task of the robot changes. In the proposed method, the training data is collected while the robot is healthy and performing some task $A$. Subsequently, the short-time Fourier transform (STFT) of the data and their combinations are calculated to derive the features used to determine the condition of the robot, i.e., these features constitute the input to the predictive maintenance algorithm. These features are in a subspace called \emph{source domain}, ${\mathcal D}_s$, representing the healthy robot. The test data is collected while the robot is doing task $B$. Test data is in the \emph{target domain}, ${\mathcal D}_t$. The goal of the transfer learning algorithm is to transfer the model learned in the source domain to the target domain~\cite{pan2010survey}. We assume that labeled data of healthy and faulty robots are not available in the target domain. Therefore, unsupervised methods must be used to transfer the knowledge between the domains. In this paper, we propose to use \emph{manifold alignment}~\cite{wang2009general, boucher2015aligning}, which is a local-preserving algorithm that finds a common subspace of the source and the target domains. Figure~\ref{fig:schematic} shows the flowchart of the proposed system.

Furthermore, since it is not convenient to ask the users of the robot to run the transfer learning algorithm each time the task of the robot changes, we assume that the test data is from the target domain. Thus, the manifold alignment algorithm is continuously applied to the features extracted from the test data to make sure that the comparison between healthy data and the test data is performed in the same domain. The experiments on real-world datasets obtained from two types of industrial robots show a significant performance improvement in dealing with operational changes, as well as the ability of the algorithm to correctly identify anomalies.
	
The following section describes the real-world dataset, pre-processing, feature extraction, and detection methods. Subsequently, the proposed predictive maintenance method based on the manifold alignment algorithm is presented in detail, followed by the presentation and discussion of experimental results based on real-world data traces. The final section concludes the findings of this paper and proposes directions for future work.   
	
\section{Scenario}\label{sec:scenario}

\subsection{Dataset}
	
	
The dataset is collected from two robot types. Robot ${\mathcal A}$ is from a series of single arm robots with 6 axis. Their reach is up to 3m (10ft) and their payload is around 100kg (220lb). Robot ${\mathcal B}$ is also from a series of 6-axis single arm robots and reach of up to 4m (13ft) and a maximum payload of 600kg (1320lb). Three signals, position, speed, and torque, are recorded from the controller of these robots for each axis separately. The sampling frequency of speed and torque signals are $2kHz$ while the position signal is sampled at a frequency of $250Hz$.
Since the dataset comprises real-world data from robots in operation, we cannot provide more details about the robots, nor their specific tasks due to confidentiality obligations. However, we believe that the generic nature of our proposed predictive maintenance method makes it possible to apply it successfully for a broad range of robot types and tasks.
	
\subsection{Pre-processing}

Since the sampling frequency of the position signal is lower than the other two signals, as mentioned above, the position signal is upsampled to get the same number of data points for each signal.
The position signal is upsampled by first inserting 7 data points with value zero between pairs of actual signals, which artificially increases the data rate to $2kHz$. Next, the zero values are overwritten using the
cubic interpolation method, which fits a third degree polynomial function to the samples of the position signal~\cite{fritsch1980monotone}. The fitted polynomial function is used to approximate the values of position signals at the newly inserted sample points. After pre-processing, each signal $x(t) = [speed(t), position(t), torque(t)]^T$ provides one sample at a time $t$, where $speed(t)$, $position(t)$, and $torque(t)$ are scalar values.

\subsection{Feature Extraction}

After the data is pre-processed, a certain set of features is required to be extracted from the given three signals. The features should not only accurately represent normal data but also discriminate normal from anomalous data well. In this paper, the short-time Fourier transform (STFT) of the three signals in $x(t)$ and their combinations is computed as the set of features. The reason for choosing STFT is that it adequately represents the time-frequency distribution of the signals, which is suitable to identify various anomalies. For example, gearbox malfunctioning and backlash appears as vibration during robot operation. STFT is able to capture these vibrations as different frequency components at different time steps. However, note that finding the appropriate features for the robot predictive maintenance is not the focus of this paper. Rather, we focus on using appropriate (given) features to detect anomalies without causing false positives due to operation changes. As we will show later, the standard STFT-based approach to identify features already yields promising results in experiments using real-world data.

\subsection{Detection}
	
The absolute values of STFT magnitude are used to build the subspace representing the healthy robot.
Naturally, the performance of the anomaly detection mechanism critically hinges upon the accurateness of this subspace.
Before delving into our proposed method to derive such a subspace in the subsequent section, we briefly summarize a common and straightforward approach: principal component analysis (PCA) can be applied to form the subspace, i.e., the principal components of the absolute STFT values constitute the subspace associated with a healthy robot.
Given this subspace, training and test datasets are projected to this subspace using the principal components. The $\ell_2$-norm distance between the training dataset (of a healthy robot) and the test dataset in this subspace is used as a criterion to identify whether the test data represents a healthy or faulty robot. If the test dataset comes from a healthy robot, the subspace should be able to represent the features of the test data well and the distance to the training data in this subspace should be small. On the other hand, a large distance is an indication of a faulty robot.

Formally, let ${\bf X}_{s}$ and ${\bf X}_{t}$ represent absolute STFT values of training (healthy) and test data, respectively,  where rows are samples and columns are the flattened absolute STFT values of a particular sample. Our distance criterion is defined as $d = \|{\bf X}_{s}{\bf P}-{\bf X}_{t}{\bf P} \|_2^2$, where ${\bf P}$ is the matrix whose columns are the principal components. We formulate the detection problem as:

\begin{equation}
    \begin{array}{cc}
         H_0: & d \ge \varepsilon \\
         H_1: & d < \varepsilon
    \end{array}
\end{equation}

\noindent where $\varepsilon$ is the threshold determining whether the test data represents a healthy or faulty condition. The parameter $\varepsilon$ can be set based on the mean distance and the variance for samples associated with a healthy robot. In other words, the threshold is determined by the significance level, $\alpha$, or the percentile of the distribution.

	
\section{Proposed Method}
\label{sec:domain}
	
The problem with existing methods is that the subspace built from the features of the training data, collected from a healthy robot, of one specific operation is not able to represent the data coming from the same healthy robot but executing another operation. As a result, the $\ell_2$-norm distance increases and the algorithm raises a false alarm. In this section, transfer learning is used to find a common subspace of the training and test data. Then, the $\ell_2$-norm distance is calculated in this common subspace.
	
Let ${\mathcal X}_s$ represent the feature space in the source domain, and ${\bf X}_s \in {\mathbb R}^{N \times K}$ are $K$ features (absolute STFT values) of $N$ training samples drawn from this space. Let ${\bf X}_t \in {\mathbb R}^{N \times K}$ be $K$ features of $N$ test samples (absolute STFT values) drawn from the target feature space ${\mathcal X}_t$ and collected while the robot is in operation. Since the application of the robot during training and test are different, their corresponding feature spaces are different, i.e., ${\mathcal X}_s \neq {\mathcal X}_t$. Therefore, there is a need for a transfer learning algorithm to reduce the difference between these two spaces while preserving the geometric properties~\cite{pan2011domain}. This reduction can be achieved by finding a common subspace between source and target spaces through minimizing a certain cost function. Manifold alignment is an unsupervised tranfer learning (or domain adaptation) algorithm that provides a closed-form solution~\cite{boucher2015aligning, wang2009general}. Having a closed-form solution makes it possible to implement it in a computationally efficient manner, which can be a requirement for applications with real-time constraints. For this reason, we chose manifold alignment to perform domain adaptation. The manifold alignment algorithms replaces the application of PCA as discussed in the previous section. Hence, the input to the manifold alignment algorithm is the STFT values, ${\bf X}_s$ and ${\bf X}_t$, and the output is the computed distance $d$.
More precisely, the manifold alignment algorithm computes low-rank embeddings (LREs) of ${\bf X}_s$ and ${\bf X}_t$ in a joint subspace. The LREs are then used to calculate the distance.

	%
	%
We will now discuss how the LREs of the source and target features, ${\bf X}_s $ and ${\bf X}_t $, are calculated.
The objective is to minimize the following loss function:
	
	\begin{equation}
		\begin{array}{c}
			\min_{R_s} {1 \over 2} {\|{\bf X}_s-{\bf X}_s{\bf R}_s \|}_F^2 + \lambda {\|{\bf R}_s\|}_* \\
			\min_{R_t} {1 \over 2} {\|{\bf X}_t-{\bf X}_t{\bf R}_t \|}_F^2 + \lambda {\|{\bf R}_t\|}_*,
		\end{array}
	\end{equation}
	
\noindent where $\lambda>0$, ${\|.\|}_F$ and ${\|.\|}_*$ are Frobenius and spectral norms, respectively. In this equation, ${\bf X}_s{\bf R}_s$ and ${\bf X}_t{\bf R}_t$ are the low rank embeddings of ${\bf X}_s $ and ${\bf X}_t $, respectively, and ${\bf R}_s$ and ${\bf R}_t$ are their reconstruction coefficient matrices. 


\begin{align}
{\bf R}_s &= {\hat {\bf V}}_s({\bf I}-{\bf \hat S}^{-2}_s){\bf \hat V}^T_s \;\;\text{and}\\
{\bf R}_t &= {\bf \hat V}_t({\bf I}-{\bf \hat S}^{-2}_t){\bf \hat V}^T_t,
\end{align}
where ${\hat {\bf S}}_s$ is the diagonal matrix of all singular values greater than one and matrix ${\hat {\bf V}}_s.$ is comprised of the corresponding right-singular vectors. The block reconstruction coefficient matrix is given by:

	\begin{equation}
		{\bf R}=\left [
		\begin{array}{cc}
			{\bf R}_s & {\bf 0} \\
			{\bf 0} & {\bf R}_t
		\end{array}
		\right ]
	\end{equation}
	
	The inter-set correspondence between the samples of the training and test datasets is represented by
	$${\bf C} = \left [
	\begin{array}{cc}
	{\bf 0} & {\bf I} \\
	{\bf I} & {\bf 0}
	\end{array}
	\right ],$$
	where ${\bf I}$ is the identity matrix. 
	
	
	After finding the LRE of source and target samples, the projection matrices from the source and the target space into the common subspace and the embedding of the source and target samples are calculated by minimizing the following cost function:
	
	\begin{equation}
		(1-\mu) \| {\bf F} - {\bf R}{\bf F} \|_F^2+\mu \sum_{i,j=1}^N{ \|{\bf F}_i-{\bf F}_j\|^2 C(i,j)},
	\end{equation} 
	
	\noindent where $\mu \in [0,1]$ determines the importance of the local geometry (first term) vs.\ the inter-set correspondence (second term). The cost function can be simplified to the following expression~\cite{boucher2015aligning}:
	
	\begin{equation}\label{eq:cost2}
		(1-\mu)({\bf I}-{\bf R})^T({\bf I}-{\bf R}) + 2\mu {\bf L},
	\end{equation}
	
\noindent where ${\bf L}$ is the Laplacian matrix of ${\bf C}$. This cost function is minimized by replacing
$$
{\bf F}=\left [ \begin{array}{c} {\bf F}_s \\ {\bf F}_t \end{array} \right ]
$$
with the $d$ smallest eigenvectors of Equation~\ref{eq:cost2}. ${\bf F}$ is the $d$-dimensional embedding of $N$ training and $N$ test features in the common subspace.
	
	Since ${\bf X}_s$ is assumed to be the training features collected when the robot is healthy, the $d$ dimensional embedding of the test dataset is compared to the embedding of the test dataset using the Euclidean distance, i.e.,
	$$\delta(t) = {\sqrt{\|{\bf F}_s - {\bf F}_t \|^2_2}}.$$
	If ${\bf X}_t$
	is anomalous, its distance is larger than for ${\bf X}_s$. In order to perform hypothesis testing on the test data, the metric $\delta$ is calculated for several normal datasets to build a probability distribution function. The empirical distribution of the metric is the positive half of the Laplace distribution with $\mu=0$, and any ${\bf X}_t$ outside of the confidence interval is marked as anomalous.


\section{Experimental Results}
\label{sec:exp}
	
In this section, the proposed algorithm is evaluated on a lab-generated dataset and two separate real-world datasets from robot ${\mathcal A}$ and robot ${\mathcal B}$.\footnote{The actual robot names cannot be provided for confidentiality reasons as the robots are used in production.} In order to show the efficacy of the proposed method in robot predictive maintenance, the standard PCA-based approach is used as well to build a subspace of the features (STFT of signals and their combinations) of the healthy robot from the training data. The extracted features of the test dataset are projected to this subspace and compared with the training dataset.

The results obtained using our proposed approach based on transfer learning are then compared against this PCA-based solution.

\begin{figure}
    \centering
    \includegraphics[width=0.47\textwidth]{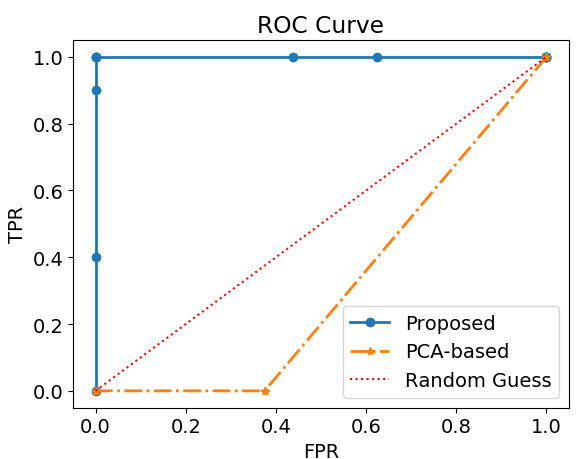}
    \caption{Receiver operating characteristic curve calculated for the $\{50, 75, 90, 95, 97.5, 99\}th$ percentiles. The experiment is repeated 10 times for each percentile and the results are averaged.}
    \label{fig:roc}
\end{figure}

\subsection{Lab Generated Dataset}

In this section, we generate a simulated dataset that resembles speed, position, and torque signals of real robots to evaluate
the effectiveness of the proposed approach. The length of each signal on each day is $N=300$ samples. The experiment spans $10$ consecutive days. The operation of the robot during the first three days does not change and the robot is healthy. The set of generated healthy signals are


\begin{flalign*}
    & Position=2 \sin(2 \pi 9 n)+\mathcal{N}(0,10^{-4})& \\
    & Speed=10 \sin(2 \pi 10 n)+\mathcal{N}(0,10^{-4}) & \\
    & Torque= \sin(2 \pi 10 n) + 0.05 \sin(2 \pi 300 n)+\mathcal{N}(0,10^{-4}),&
\end{flalign*}

\noindent where $\mathcal{N}(0,10^{-4})$ is an added noise randomly generated from a Gaussian distribution with zero mean and standard deviation of $10^{-4}$. The robot arm, in a typical operation, moves mostly between two locations (pick and place). Therefore, the position signal looks like a combination of sinus signals oscillating over time. Since the speed and torque have a direct relationship with the position signal, they also resemble combinations of sinus signals. However, the torque signal is typically noisy and has a saw-edged shape. For this reason, we  added the term $0.05 \sin(2 \pi 300 n)$, which results in torque signals that are similar in shape to those observed in real-world applications.     

On the fourth day, the operation does not change---which means the three signals will be the same as before. However, the torque signal shows an anomaly, which starts at the $150^\mathit{th}$ sample and lasts for $60$ samples. For these $60$ samples, the torque signal is


\begin{alignat*}{1}
    Torque &= \sin(2 \pi 10 n) + 0.3 \sin(2 \pi 100 n) \\
    & + 0.3 \sin(2 \pi  n)+\mathcal{N}(0,10^{-4}).
\end{alignat*}

The robot continues its operation one more day with normal behavior and the same operation. On the sixth day, the operation of the robot changes to

\begin{flalign*}
    & Position=2 \sin(2 \pi 4 n)+\mathcal{N}(0,1e^{-4})& \\
    & Speed=10 \sin(2 \pi 5 n)+\mathcal{N}(0,1e^{-4}) & \\
    & Torque= \sin(2 \pi 5 n) + 0.05 \sin(2 \pi 300 n)+\mathcal{N}(0,1e^{-4}),&
\end{flalign*}

\noindent and lasts for three days. Then, the operation goes back to the first operation. Thus, the total $10$ days of operation can be summarized as $\{O_1, O_1, O_1, A_{O_1}, O_1, O_2, O_2, O_2, O_1 \}$, where $O_1$ and $O_2$ denote the first and second operations, respectively, and $A_{O_1}$ is the anomaly happening while the robot is performing operation $O_1$. A correct algorithm detects only this single anomaly. Detecting any operation change is considered as a false alarm.

We have applied the proposed algorithm to this generated dataset and calculated the true positive rate (TPR) and false positive rate (FPR) for the $\{50, 75, 90, 95, 97.5, 99\}th$ percentiles. This experiment was repeated $10$ times for both the proposed algorithm and the PCA-based algorithm at each percentile level and the results are averaged. Figure~\ref{fig:roc} shows the receiver operating characteristic (ROC) curve. As the figure shows, the ROC curve of the PCA-based method falls below random guess line while the proposed algorithm has the best possible shape, which proves the efficacy of the proposed method. 


	\begin{figure*}[t]
	    \centering
		\begin{minipage}[b]{.49\textwidth}
			\centering
			\centerline{\includegraphics[width=8cm,height=4.5cm]{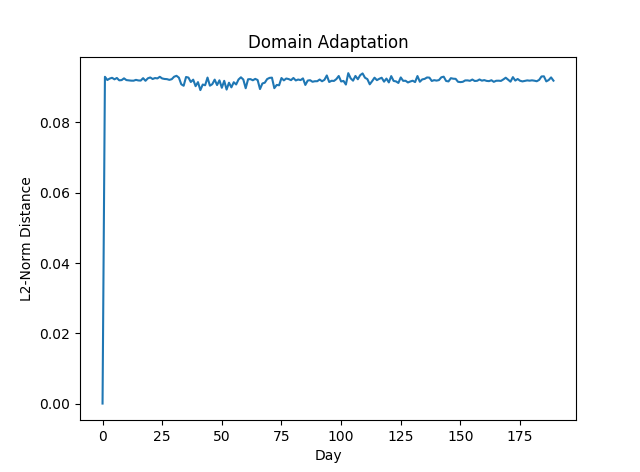}}
			\centerline{   (a)}\medskip
			\vspace{-2mm}
		\end{minipage} %
		\begin{minipage}[b]{.49\textwidth}
			\centering
			\centerline{\includegraphics[width=8cm,height=4.4cm]{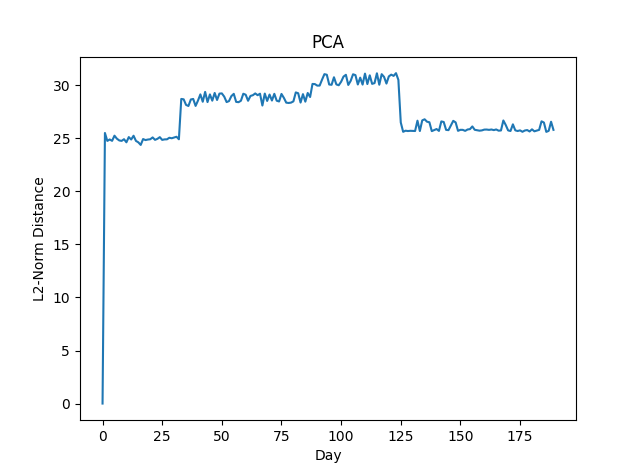}}
			\centerline{   (b)}\medskip
		\end{minipage}
		\caption{{Robot ${\mathcal A}$}: first experiment using a 190-day data trace without any anomalies and three changes in the task of the robot. (a) Distance values over 190 days calculated using the proposed method. (b) Distance values over 190 days calculated using the PCA-based method. Three changes in the task are apparent on the $30^{\mathit{th}}$, $87^{\mathit{th}}$, and $125^{\mathit{th}}$ day in (b), whereas there is no discernible change in the distance in (a).}
		\vspace{-2mm}
		\label{fig:exp1}
	\end{figure*}

	\begin{figure*}[t]
	    \centering
		\begin{minipage}[b]{0.49\linewidth}
			\centering
			\centerline{\includegraphics[width=8cm,height=4.5cm]{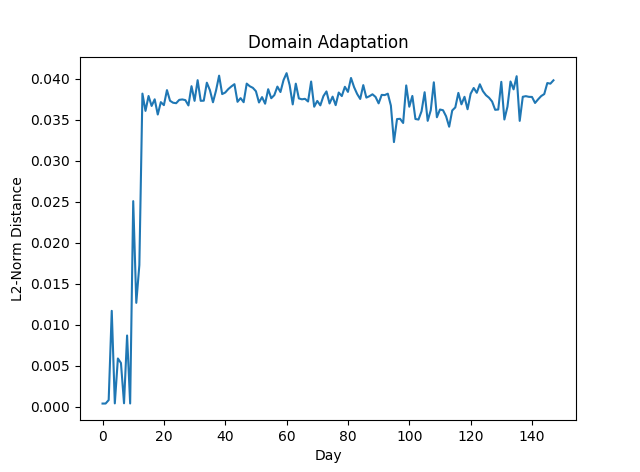}}
			\centerline{(a)}\medskip
			\vspace{-2mm}
		\end{minipage}%
		\begin{minipage}[b]{0.49\linewidth}
			\centering
			\centerline{\includegraphics[width=8cm,height=4.5cm]{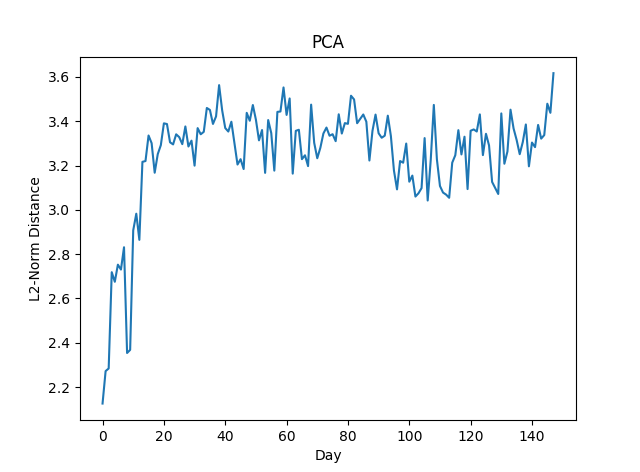}}
			\centerline{(b)}\medskip
			\vspace{-2mm}
		\end{minipage}
		\caption{{Robot ${\mathcal B}$}: second experiment using a 148-day data trace with an anomaly on the last day and no change in the task of the robot. (a) Distance values over 148 days calculated using the proposed method. (b) Distance values over 148 days calculated using the PCA-based method. The anomaly is apparent on the last day in both (a) and (b) but the relative change is more pronounced using the proposed method.}
		\label{fig:exp2}
	\end{figure*}

\subsection{Real-World Dataset}

\subsubsection{Robot ${\mathcal A}$}
	
Three signals (speed, position, torque) of the fourth axis of robot ${\mathcal A}$ were collected for 190 consecutive days. The dataset on each day is 3 seconds long. No break down or anomaly was reported for this axis of the robot; however, the task of the robot changed three times during these 190 days. After pre-processing, the dataset on day 1 is used as the training data (healthy robot) and the respective STFT values constitute the extracted features ${\bf X}_s$. The datasets of the following days are used as the test dataset to form ${\bf X}_t$. $\delta(t)$ is calculated for each day with respect to the first day to identify any changes with respect to the first day. The assumption is that the robot is healthy on the first day, which can be the time of inspection or commissioning at the robot manufacturing facility. Figure~\ref{fig:exp1}(a) shows $\delta(t), \hspace{2mm} t \in \{1,2,\ldots,190\}$. It is apparent in this figure that none of the changes in the task of the robot is identified as anomalous behavior when using our proposed method based on transfer learning (domain adaption). On the other hand, Figure~\ref{fig:exp1}(b) shows a different picture when using PCA projection. All three changes in the task of the robot are obvious in this figure. The resulting changes in the computed distances indicate that the use of conventional methods such as PCA fails to distinguish between a change in the mechanical condition of the robot from a change in its task. A second observation when comparing these two figures is the magnitude of the distance function $\delta(t)$. The $\delta(t)$ values in Figure~\ref{fig:exp1}(b) are significantly larger than those of Figure~\ref{fig:exp1}(a). The larger distances provide further evidence that ${\bf X}_s$ and ${\bf X}_t$s are not in the same subspace. On the other hand, the small values of $\delta(t)$ in Figure~\ref{fig:exp1}(a) indicate that their projection onto the common subspace renders them comparable.
	
\subsubsection{Robot ${\mathcal B}$}
In the second experiment, three signals of the fifth axis of robot ${\mathcal B}$ were collected for 148 days. The dataset on each day is again 3 seconds long as in the first experiment. The axis of the robot broke down on the $149^{\mathit{th}}$ day. However, the task of the robot never changed during the entire 148 days. The training and test data sets are constructed as before:
The first day is considered the training data and the features are extracted from this dataset as ${\bf X}_s$. The extracted features of the other $147$ days form ${\bf X}_t$. The test data is compared to the training data using the $\ell_2$-norm to obtain the distance $\delta(t)$ at time $t$. Figure~\ref{fig:exp2}(a) shows the computed distances of the proposed algorithm. As in the first experiment, PCA is also used to build the subspace of the healthy robot, and the resulting distance function is depicted in Figure~\ref{fig:exp2}(b). According to the two plots in Figure~\ref{fig:exp2}, both methods perform equally well in identifying the trend of $\delta(t)$ leading to the breakdown of the axis on the $149^{\mathit{th}}$ day. However, comparing the $\delta(t)$ values of both plots demonstrates the training and test data are being compared in the same subspace in the proposed algorithm in contrast to the PCA-based method in Figure~\ref{fig:exp2}(b). As a result, there is a significantly sharper relative increase in the distance around and on the day the anomaly occurred when using our proposed method, implying that it can more distinctly identify anomalies.
	
\section{Conclusion}
	
Predictive maintenance systems are of great interest to  manufacturers of various kinds of mechanical devices and apparatuses such as drives and robots. The goal of such systems is to build a model of the healthy apparatus and continuously compare the current condition of the device, captured in the form of measured signals, with the model to identify any possible errors or  malfunction. The challenge with the application of predictive maintenance systems in robotics is that the model requires to be retrained every time the task of the robot changes using conventional methods. We showed that transfer learning (domain adaptation) addresses this challenge. While we used manifold alignment algorithm to project the features extracted from the training and test data onto a common subspace, it is worth investigating other unsupervised transfer learning algorithms. We used distance-based methods to compare the test data with the training data in the common subspace to identify any anomaly. The future work will consider time-series analysis methods in the common subspace to capture the dynamics of systems.

	
	
	%
	%
	
	\bibliographystyle{IEEEbib}
	\bibliography{strings}

\begin{thebibliography}{10}

\bibitem{hornung2014model}
Rachel Hornung, Holger Urbanek, Julian Klodmann, Christian Osendorfer, and
  Patrick Van Der~Smagt,
\newblock ``Model-free robot anomaly detection,''
\newblock in {\em Intelligent Robots and Systems (IROS 2014), 2014 IEEE/RSJ
  International Conference on}. IEEE, 2014, pp. 3676--3683.

\bibitem{liu2005model}
Honghai Liu and George~M. Coghill,
\newblock ``A model-based approach to robot fault diagnosis,''
\newblock {\em Knowledge-Based Systems}, vol. 18, no. 4-5, pp. 225--233, 2005.

\bibitem{mobley2002introduction}
R~Keith Mobley,
\newblock {\em An introduction to predictive maintenance},
\newblock Elsevier, 2002.

\bibitem{garcia2006simap}
Mari~Cruz Garcia, Miguel~A Sanz-Bobi, and Javier Del~Pico,
\newblock ``Simap: Intelligent system for predictive maintenance: Application
  to the health condition monitoring of a windturbine gearbox,''
\newblock {\em Computers in Industry}, vol. 57, no. 6, pp. 552--568, 2006.

\bibitem{grall2002continuous}
Antoine Grall, Laurence Dieulle, Christophe B{\'e}renguer, and Michel
  Roussignol,
\newblock ``Continuous-time predictive-maintenance scheduling for a
  deteriorating system,''
\newblock {\em IEEE transactions on reliability}, vol. 51, no. 2, pp. 141--150,
  2002.

\bibitem{hashemian2010state}
Hashem~M Hashemian,
\newblock ``State-of-the-art predictive maintenance techniques,''
\newblock {\em IEEE Transactions on Instrumentation and measurement}, vol. 60,
  no. 1, pp. 226--236, 2010.

\bibitem{park2016multimodal}
Daehyung Park, Zackory Erickson, Tapomayukh Bhattacharjee, and Charles Kemp,
\newblock ``Multimodal execution monitoring for anomaly detection during robot
  manipulation,''
\newblock in {\em Robotics and Automation (ICRA), 2016 IEEE International
  Conference on}. IEEE, 2016, pp. 407--414.

\bibitem{borgi2017data}
Tawfik Borgi, Adel Hidri, Benjamin Neef, and Mohamed~Saber Naceur,
\newblock ``Data analytics for predictive maintenance of industrial robots,''
\newblock in {\em 2017 International Conference on Advanced Systems and
  Electric Technologies (IC\_ASET)}. IEEE, 2017, pp. 412--417.

\bibitem{sjostrand2012dynamic}
Niclas Sj{\"o}strand, Dominique Blanc, and Anders Lindin,
\newblock ``Dynamic maintenance plan for an industrial robot,'' May~22 2012,
\newblock US Patent 8,185,346.

\bibitem{pinto2019robot}
Riccardo Pinto and Tania Cerquitelli,
\newblock ``Robot fault detection and remaining life estimation for predictive
  maintenance,''
\newblock {\em Procedia Computer Science}, vol. 151, pp. 709--716, 2019.

\bibitem{moisescu2018retracted}
Mihnea~Alexandru Moisescu, Ioan~Stefan Sacala, Ioan Dumitrache, and Simona
  Caramihai,
\newblock ``Retracted: Predictive maintenance and robotic system design,''
\newblock {\em Journal of Fundamental and Applied Sciences}, vol. 10, no. 4S,
  pp. 234--239, 2018.

\bibitem{sathish2019data}
Vallachira Sathish, Michal Orkisz, Mikael Norrlof, and Sachit Butail,
\newblock ``Data-driven gearbox failure detection in industrial robots,''
\newblock {\em IEEE Transactions on Industrial Informatics}, 2019.

\bibitem{pan2010survey}
Sinno~Jialin Pan and Qiang Yang,
\newblock ``A survey on transfer learning,''
\newblock {\em IEEE Transactions on knowledge and data engineering}, vol. 22,
  no. 10, pp. 1345--1359, 2010.

\bibitem{wang2009general}
Chang Wang and Sridhar Mahadevan,
\newblock ``A general framework for manifold alignment,''
\newblock in {\em AAAI Fall Symposium: Manifold Learning and Its Applications},
  2009.

\bibitem{boucher2015aligning}
Thomas Boucher, Cj~Carey, Sridhar Mahadevan, and Melinda~Darby Dyar,
\newblock ``Aligning mixed manifolds.,''
\newblock in {\em AAAI}, 2015, pp. 2511--2517.

\bibitem{fritsch1980monotone}
Frederick~N Fritsch and Ralph~E Carlson,
\newblock ``Monotone piecewise cubic interpolation,''
\newblock {\em SIAM Journal on Numerical Analysis}, vol. 17, no. 2, pp.
  238--246, 1980.

\bibitem{pan2011domain}
Sinno~Jialin. Pan, Ivor~W. Tsang, James~T. Kwok, and Qiang Yang,
\newblock ``Domain adaptation via transfer component analysis,''
\newblock {\em IEEE Transactions on Neural Networks}, vol. 22, no. 2, pp.
  199--210, 2011.

\end{thebibliography}

\end{document}